\documentclass{article}

\usepackage{arxiv}

\usepackage[utf8]{inputenc} 
\usepackage[T1]{fontenc}    
\usepackage{hyperref}       
\usepackage{url}            
\usepackage{booktabs}       
\usepackage{amsfonts}       
\usepackage{nicefrac}       
\usepackage{microtype}      
\usepackage{lipsum}
\usepackage{graphicx}
\graphicspath{ {./images/} }
\usepackage[export]{adjustbox}

\title{mAedesID: Android Application for Aedes Mosquito Species Identification using Convolutional Neural Network}

\author{
 G. Jeyakodi \\
  Department of Computer Science\\
  Pondicherry University\\
  Puducherry, India \\
  \texttt{rjeyakodi02@gmail.com} \\
   \And
 Trisha Agarwal \\
  Department of Computer Science\\
  Pondicherry University\\
  Puducherry, India \\
  \texttt{trisha.agarwal27@gmail.com} \\
  \And
 P. Shanthi Bala \\
  Department of Computer Science\\
  Pondicherry University\\
  Puducherry, India \\
  \texttt{shanthibala.cs@gmail.com} \\
}

\begin{document}
\maketitle
\begin{abstract}
Vector-Borne Disease (VBD) is an infectious disease transmitted through the pathogenic female Aedes mosquito to humans and animals. It is important to control dengue disease by reducing the spread of Aedes mosquito vectors.  Community awareness plays a crucial role to ensure Aedes control programmes and encourages the communities to involve active participation. Identifying the species of mosquito will help to recognize the mosquito density in the locality and intensifying mosquito control efforts in particular areas. This will help in avoiding Aedes breeding sites around residential areas and reduce adult mosquitoes. To serve this purpose, an android application are developed to identify Aedes species that help the community to contribute in mosquito control events. Several Android applications have been developed to identify species like birds, plant species, and Anopheles mosquito species. In this work, a user-friendly mobile application ‘mAedesID’ is developed for identifying the Aedes mosquito species using a deep learning Convolutional Neural Network (CNN)  algorithm which is best suited for species image classification and achieves better accuracy for voluminous images. 

\end{abstract}


\section{Introduction}
World Health Organization (WHO) stated that dengue is considered one of the ten high-priority diseases that pose global public health threats. The dengue virus is recognized as a pathogen for the  Aedes genus mosquitoes Aedes aegypti and Aedes albopictus. These mosquitoes transfer the dengue virus to humans and it causes dengue disease. Throughout the world, these mosquitoes are documented as primary and secondary vectors. The distribution of these species also keeps on the increase due to globalization and international travel. There is an enormous increase in dengue cases in India during the past ten years and worldwide it was reported around 125 tropical and subtropical countries \cite{r1,r2}. Still, now there is no standard vaccination is available for dengue, the only way to control dengue is through vector control measures. Several automated methods are introduced to identify reliable mosquito species. Some of the examples are the wingbeat frequency analysis, computer vision approaches, and deep Convolutional Neural Networks to predict the appropriate mosquito types using images \cite{r3,r4,r5,r6}. It is necessary to develop an Aedes species image identification model that can easily run on android based mobile devices with low latent.

There are only limited mobile applications existing related to Aedes genus mosquitoes. The information based on the study, entertainment, alertness, and helpline provisions is explored. The awareness of Aedes species biology, their pathogenic behavior, the disease it spreads, the prevention mechanism, and control measures need to be explored for public access.  The major requirement is the complaint notification registration on Aedes species abundance place once it is identified, to the concerned control unit for taking necessary actions for clearing the breeding places of Aedes species. Only genuine complaints have to be accepted by the application. This necessitates the requirement of a mobile application for public participation in Aedes mosquito control by identifying and preventing mosquito abundance areas.

DISapp and MOSapp are mobile applications developed in India for collecting and uploading surveillance disease data from the public and field workers \cite{r7}. The dengue awareness mobile application has been launched by the Tamil Nadu government to provide information on awareness, causes, and prevention. However, the community involvement in Aedes mosquito control is not included \cite{r8}. The mobile Short Message Service (SMS) has been introduced in Nepal for providing dengue prevention details and improving dengue control practices \cite{r9}. In Fiji, the mobile application using Global Positioning System (GPS) technology identifies the dengue abundance areas \cite{r10}. The Convolutional Neural Network-based deep learning method is proved to be more effective as compared with the conventional feature-based method for classifying mosquito species \cite{r5}. J Pablo et. al (2018) developed a system to identify disease-carrying insects based on computer vision technology to facilitate the communities in arbovirus epidemics \cite{r11}. 

Analysis of the different computer-aided technologies to identify skin disease shows that deep learning-based image recognition methods give better results \cite{r12}. A pictographic android application is more supportive for entomologists, researchers, and public health workers \cite{r13}. The performance evaluation of CNN architectures revealed that MobileNetV2 architecture achieved the best on the verified metrics \cite{r14}. An automated system for treating and predicting the early stages of Chagas disease has been developed for healthcare clinicians which can completely cure Chagas disease \cite{r15}. Mobile technology has been adapted in ICMR-National Institute for Research in Tribal Health, Jabalpur for dengue disease diagnosis in a prior stage and promoting the mosquito population reduction activity \cite{r16}. In Sri Lanka, a mobile application has been used for educating the dengue disease to the community, school children, and field workers. The facility for reporting dengue incidents to Public Health Inspectors is provided for helping them to analyze the dengue case reports \cite{r17}. The Aedes mosquito prevention and monitoring technology have been studied by Geovanna Cristine de Souza Silva et al (2018) for building a model for Aedes control \cite{r18}.

The advancement in different Neural Network algorithms can be used to detect the type of mosquito species without having any morphological characteristics, microscope, and polymerase chain reaction (PCR) test. The image captured by a camera is sufficient to recognize the mosquito genus or species type quickly using artificial intelligence algorithms. A model that uses a deep learning image classification technique can assist the communities in detecting life-threatening Aedes mosquito species. In this work mAedesID, an android application is developed for identifying Ae. aegypti and Ae. albopictus species at the community level. This application is based on a sequential CNN model that identifies certain features and uses them to differentiate between different images and assign labels to them. The model accepts an input image and produces a prediction for the class which it belongs to based on the training. It can also be extended for identifying different mosquito genera such as Anopheles, Culex, and Mansonia by replacing the existing data with new data set and training the model accordingly. The model can also be extended for other image classification problems in which the mobile application can be developed.

\section{Methodology}
The methodology design consists of various phases such as Data Collection, Data Preprocessing, Model Building, Model Evaluation, Android Application Development, and Application Testing as depicted in \textbf{Fig 1}.

\begin{figure} 
    \centering
    \includegraphics[width=\textwidth]{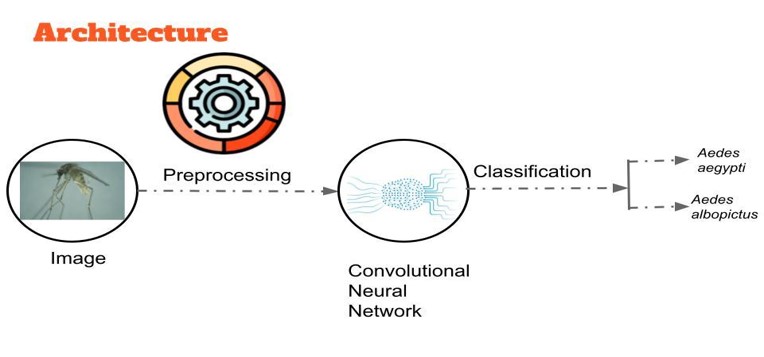}
    \caption{Methodology Design.}
\end{figure}

\label{sec:headings}

\paragraph{Data Collection.}
The input data is the collection of images of Aedes genus mosquito species Ae. aegypti and Ae. albopictus. The species images were downloaded from the Kaggle repository \cite{r19}. The 4803 Aedes images of 20.36 GB were downloaded. The device Aedes Detector was used to capture the images in different angles with high resolution. 

\paragraph{Data Preprocessing.}
IThe downloaded images were preprocessed to improve the clarity, highlight the features, and structured the images. The zero-phase component analysis (ZCA) is performed on the dataset to reduce redundancy in the matrix of pixel images. Images were normalized to ensure that all the images contribute more evenly to the total loss and process the input faster. Images were rescaled to 1/255 for converting the pixels from the range [0,255] to  [0,1]. Python tensor flow’s Keras package was used for image preprocessing. \textbf{Fig 2} depicts the data preprocessing code.
\begin{figure} 
    \centering
    \includegraphics[width=\textwidth]{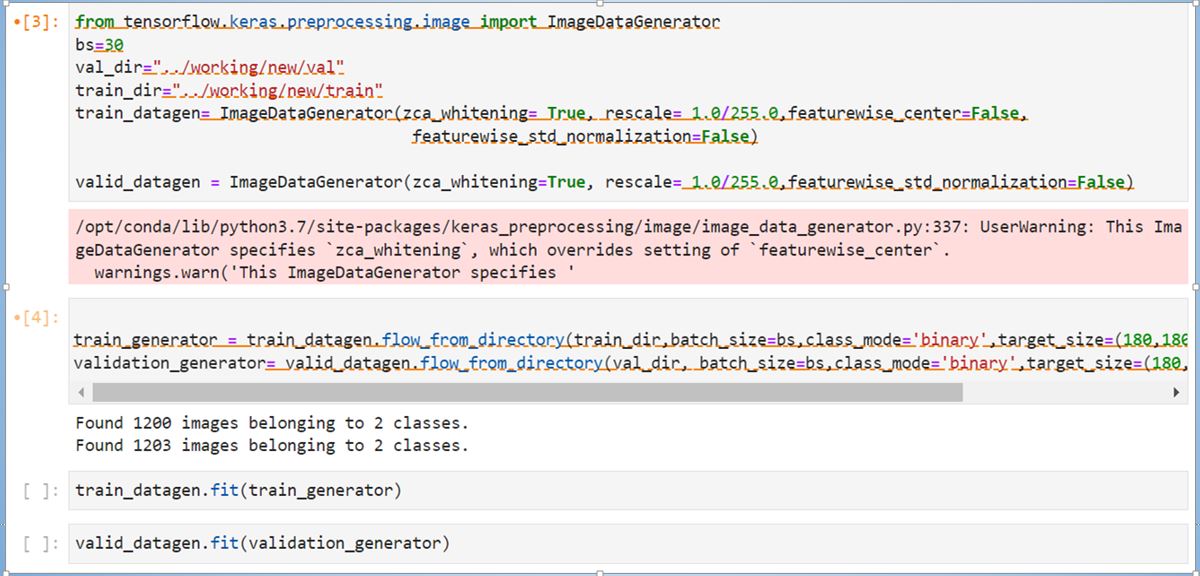}
    \caption{Code for Image Preprocessing.}
  \end{figure}

\paragraph{Model Building.}
In this phase, a sequential model has been built using Convolutional Neural Network for processing the training data. In a sequential model, multiple layers are stacked such as the output of the previous layer is the input of the new layer. CNN consists of an input layer, hidden layer, and output layer. The input layer is the first layer of the sequential model where the input image is shaped as (180,180,3) since the input images are of (180,180) pixels. Three refers to the Red Green Blue (RGB) channel implying the colored images. After processing the incoming data, it is distributed to the hidden layers that comprise the convolutional layers, major building elements, dense layers, dropout layers, flatten layers, and max-pooling layers. The dense layer connects the neural network layer deeply and the dropout layer is used to prevent the model from overfitting. Flatten layer transforms a 2D matrix of features into a vector of features. The last dense layer is the output layer that uses the sigmoid activation function x is equal to, 
\begin{equation}
\sigma(x) =  1/(1 + exp(-x))
\end{equation}

 \textbf{Fig 3} represents the code for the CNN model that is built-in Google Cloud Artificial Intelligence (AI) platform through Jupiter Notebook. \textbf{Fig 4} shows the code for model compiling and fitting using 30 epochs. 

\begin{figure} 
    \centering
    \includegraphics[width=\textwidth]{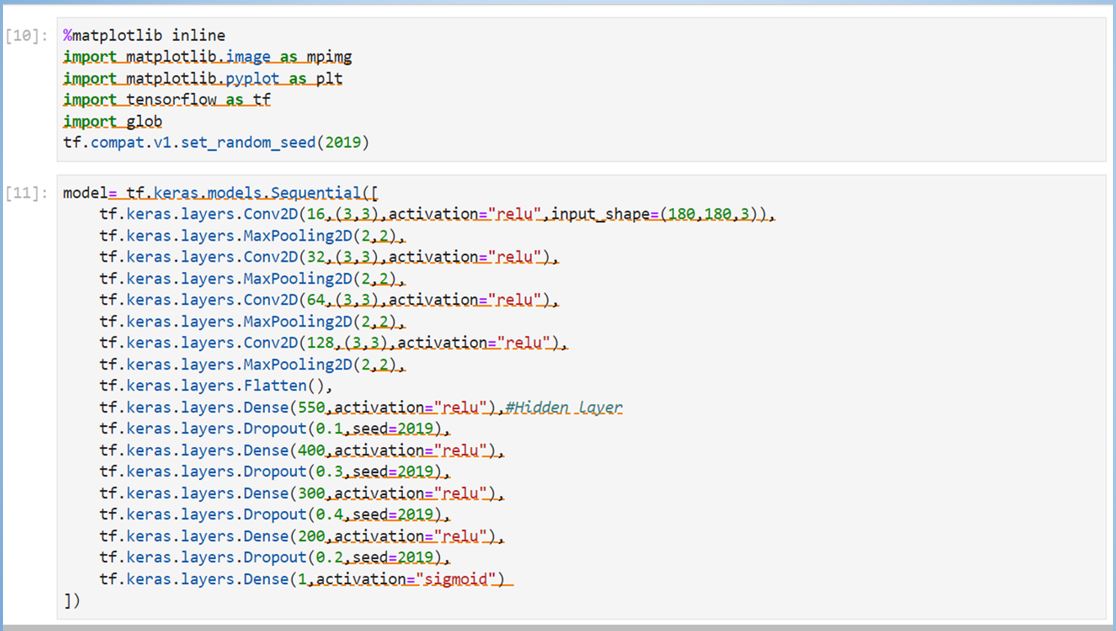}
    \caption{Code for CNN Model Building.}
\end{figure}
 
\begin{figure} 
    \centering
    \includegraphics[width=\textwidth]{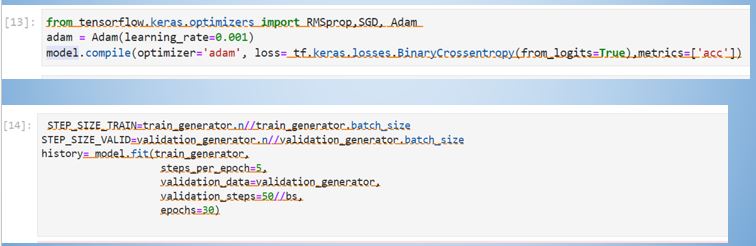}
    \caption{Code for Image Preprocessing.}
\end{figure}

\paragraph{Model Evaluation.}
For better model fitting 30 epochs were used and received two metrics for each epoch acc and val\_acc showing the accuracy in prediction obtained from training and validation data respectively. The accuracy of the model is further improved by using appropriate library functions. The developed TensorFlow model helps to predict the images on test data. \textbf{Fig 5} shows the code for testing data predictions.

\begin{figure} 
    \centering
    \includegraphics[width=\textwidth]{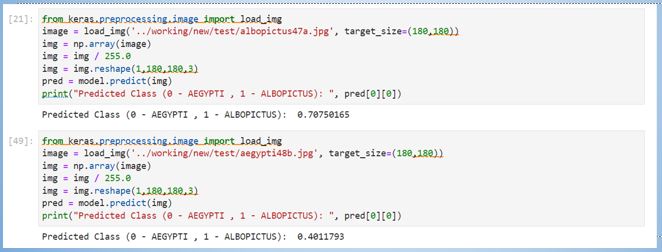}
    \caption{Code for Test Data Prediction.}
\end{figure}

\paragraph{Android Application Development.}
The TensorFlow model developed in the Google Cloud AI platform is converted into Tensorflow Lite flat buffer file (.tflite) using Tensorflow Lite Converter for Firebase storage. mAedesID, the android application is developed using android studio in Java. The mobile application identifies the Aedes species type based on the sigmoid value. \textbf{Fig 6} shows the use case diagram of mAedesID development. The user can upload either the stored image or capture the image using a mobile camera for species identification.
\begin{figure} 
    \centering
    \includegraphics[width=\textwidth]{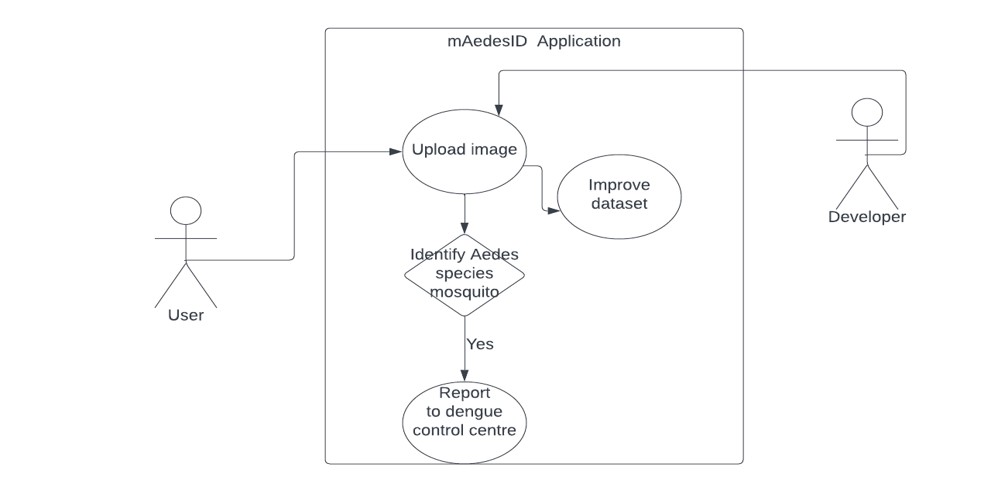}
    \caption{Use case Diagram of mAedesID.}
\end{figure}

\paragraph{Application Testing.}
The compatibility testing and functional testing were done on the mobile application. The developed mobile application has been tested with various mobile devices to check whether the mobile application user interface is compatible with all kinds of screen sizes and different kinds of operating systems (OS). It has been found that this mobile application has worked properly in all kinds of screen sizes and android OS versions 6.0 and above. The developed mobile application is tested to check its functionality is working properly or if any problems or issues were raised. But there was no issue and all the functions worked effectively.

\section{Results and Discussion}
Kaggle dataset has been used to build an image classification model using Convolutional Neural Network to identify the Aedes mosquito species through an android application. \textbf{Fig 7} shows the input image dataset that are downloaded from Kaggle. The model was constructed with one input layer, sixteen hidden layers, and one output layer. \textbf{Fig 8} shows the summary of the CNN sequential model. The training and validation data was executed for 30 epochs to improve the accuracy. \textbf{Fig 9} shows the model fitting accuracy with the accuracy parameters. The model accuracy has been increased from 68.5\% to 84.87\% using the evaluate\_generator function as shown in \textbf{Fig 10}. \textbf{Fig 11} shows the code for loading the \.tflite file and getting predictions for raw \(untested\) data. The mobile application has been developed using android studio. The user can detect the Aedes species type by uploading the image. The sigmoid value nearby zero predicted the image as Ae. aegypti and nearby one is identified as Ae. albopictus. \textbf{Fig 12} shows the front page of mobile application and the classified result of the test images Ae. aegypti and Ae. albopictus.

\begin{figure} 
    \centering
    \includegraphics[width=\textwidth]{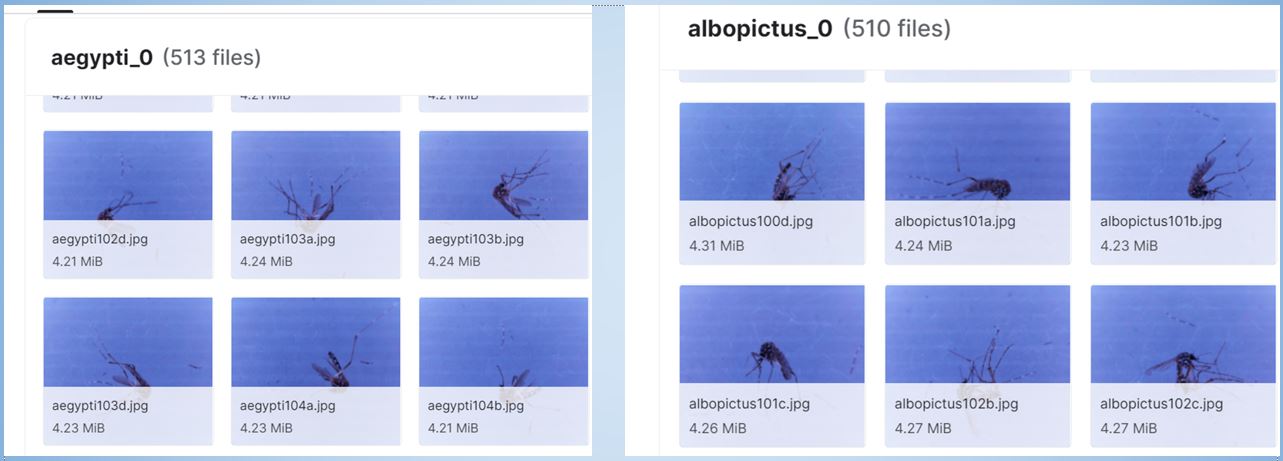}
    \caption{Input Images Dataset.}
\end{figure}
\begin{figure} 
    \centering
    \includegraphics[width=\textwidth]{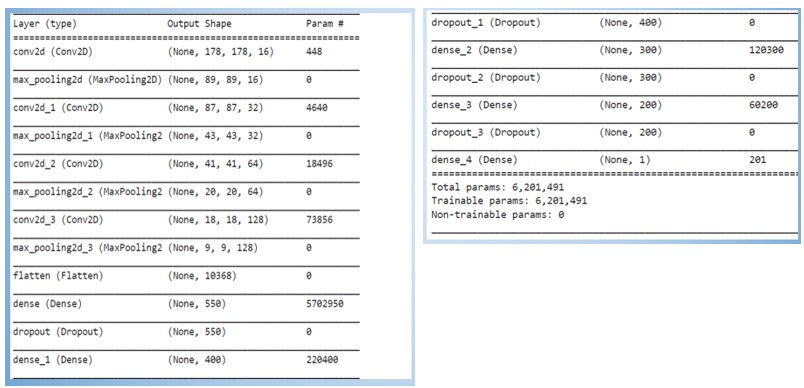}
    \caption{Classification Model Summary.}
\end{figure}
\begin{figure} 
    \centering
    \includegraphics[width=\textwidth]{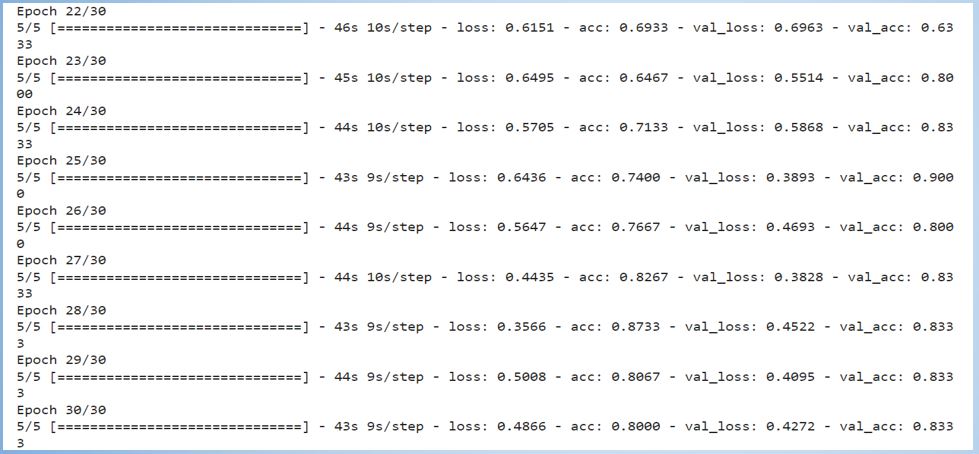}
    \caption{Model Fitting Accuracy.}
\end{figure}
\begin{figure} 
    \centering
    \includegraphics[width=\textwidth]{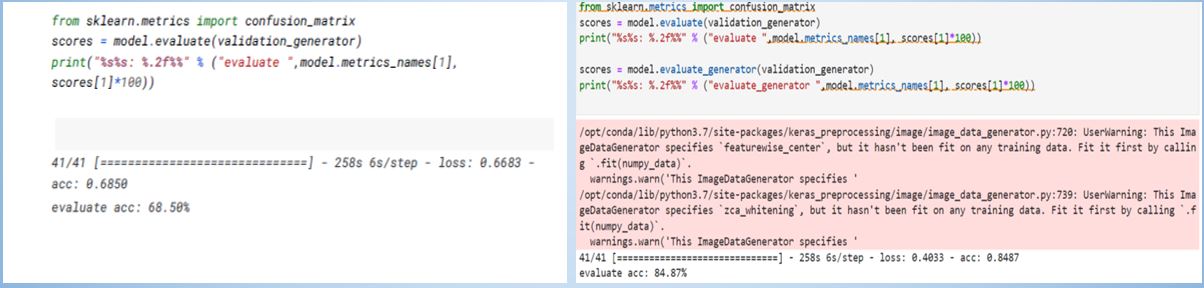}
    \caption{Accuracy of the Model}
\end{figure}
\begin{figure} 
    \centering
    \includegraphics[width=\textwidth]{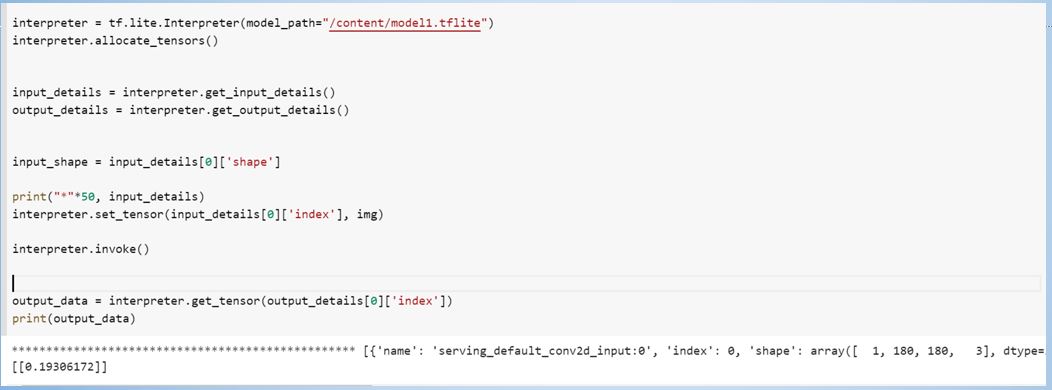}
    \caption{Code for Class Prediction on New Data.}
\end{figure}\begin{figure} 
    \centering
    \includegraphics[width=\textwidth]{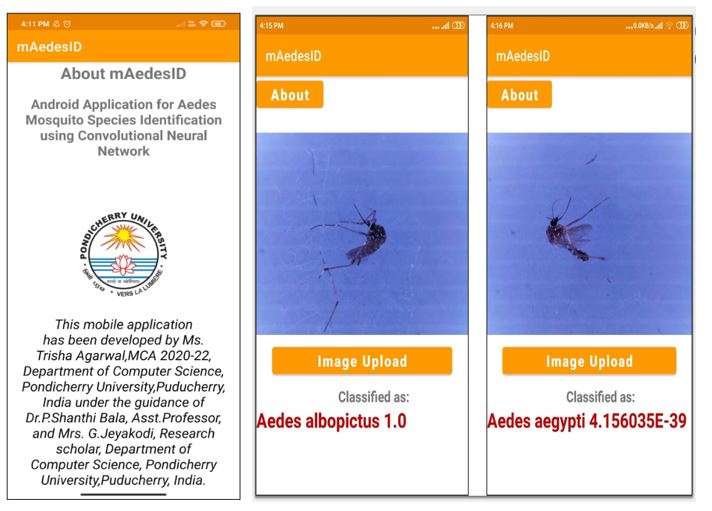}
    \caption{mAedesID – Front Page and Test Image Classified Result.}
\end{figure}

\section{Conclusion}
The mAedesID, the android application for identifying the mosquito vector Aedes aegypti and Aedes albopictus that is responsible for dengue disease was developed using the Convolutional Neural Network image classification model.  The model was built and examined with Kaggle repository Aedes species images and produced valid predictions with good accuracy. Since the model is trained with the high-quality images obtained from the Aedes Detector device, the mobile application requires the same kind of images for Aedes species classification. The mAedesID is user-friendly and helps the community to involve active participation in dengue mosquito control without any background information on the mosquito's physical appearance. In the future, the model will be extended for predicting the mobile captured Aedes images by including them in the training data and labeling unknown images. Currently, the user can download the mobile application from the URL link, after including additional functionalities it will be uploaded to the google play store for easy access. The mobile application will be the resource for dengue disease prevalence and control in dengue epidemic areas.

\bibliographystyle{unsrt}  
\bibliography{mAedesID}

\end{document}